Department of Information Management

National Yunlin University of Science & Technology

Master Thesis

Using Super-Resolution Imaging for Recognition of Low-Resolution Blurred License Plates: A Comparative Study of Real-ESRGAN, A-ESRGAN, and StarSRGAN.

Ching-Hsiang, Wang

Advisor: Hsueh-Chi Shih, Ph.D.

June 2024

# Abstract

With the robust development of technology, license plate recognition technology can now be properly applied in various scenarios, such as road monitoring, tracking of stolen vehicles, detection at parking lot entrances and exits, and so on. However, the precondition for these applications to function normally is that the license plate must be 'clear' enough to be recognized by the system with the correct license plate number. If the license plate becomes blurred due to some external factors, then the accuracy of recognition will be greatly reduced. Although there are many road surveillance cameras in Taiwan, the quality of most cameras is not good, often leading to the inability to recognize license plate numbers due to low photo resolution. Therefore, this study focuses on using super-resolution technology to process blurred license plates. This study will mainly fine-tune three super-resolution models: Real-ESRGAN, A-ESRGAN, and StarSRGAN, and compare their effectiveness in enhancing the resolution of license plate photos and enabling accurate license plate recognition. By comparing different super-resolution models, it is hoped to find the most suitable model for this task, providing valuable references for future researchers.

**Key words**：Deep Learning, Super-Resolution, License Plate Recognition, Blurry License Plates

# Table of Contents

# List of Figures

# List of Tables

# 1. Introduction

## 1.1 Research Background and Motivation

License plate recognition is a significant technology with wide applications in modern society. With increasing urban traffic congestion and road safety demands, license plate recognition technology has become a crucial tool in various fields such as law enforcement, parking management, toll systems, and security surveillance.

Research in this area has been underway for years, with its applications continually expanding. For example, parking lots are now using license plate recognition technology to save costs and increase efficiency[1]. Researchers have also focused on recognizing various types of license plates in China [2], and the system designed in[3] can accurately identify license plate numbers from different countries. However, there are still many issues and limitations with this technology that need to be addressed.

In public settings, surveillance cameras are essential tools for the functioning of license plate recognition. According to a 2021 IHS Markit report on global surveillance camera penetration, Taiwan ranks third worldwide, with an average of one surveillance camera per 5.5 people, trailing only behind the United States (one camera per 4.6 people) and China (one camera per 4.1 people). Despite these numbers, there are still issues troubling Taiwanese society.

According to statistics from Taiwan's Road Traffic Safety Commission, the number of traffic accidents has been steadily increasing each year, rising from 170,127 cases in 2008 to 375,844 cases in 2022. Hit-and-run incidents account for approximately 4.5% of these cases. For victims of hit-and-run accidents, surveillance cameras are often the only source of information about the perpetrator. However, the quality of surveillance cameras at intersections in Taiwan varies, particularly in rural areas, where the footage is often too blurry to clearly discern license plate numbers.

## 1.2 Research Objective

References[4] and [5] both mention that despite significant improvements in current license plate recognition technology, there is still room for enhancement in complex environments. They highlight many challenges that future researchers need to address, one of which is the issue of blurry license plates. Therefore, this study primarily focuses on fine-tuning three super-resolution models: Real-ESRGAN, A-ESRGAN, and StarSRGAN. It aims to compare the effectiveness of these models in enhancing the resolution of license plate photos and ensuring accurate license plate recognition. By comparing different super-resolution models, the study hopes to identify the most suitable model for this task, providing valuable references for future researchers.

## 1.3 Thesis Structure Diagram

This research paper is divided into five main chapters. Chapter 1 is the introduction, where the research background, motivation, and objectives are presented. Chapter 2 focuses on literature review, revisiting past related studies, and explaining the models used in this research. Chapter 3 delves into the research methodology, detailing data sources, preprocessing steps, and more. This chapter also introduces model evaluation methods and the experimental environment. Chapter 4 presents experimental results and analysis, discussing and analyzing the experimental design and outcomes. Finally, Chapter 5 concludes the paper, summarizing the findings based on the results, identifying limitations and areas for improvement in this study, and suggesting potential areas for future research. The structure of the thesis is illustrated in Figure 1.

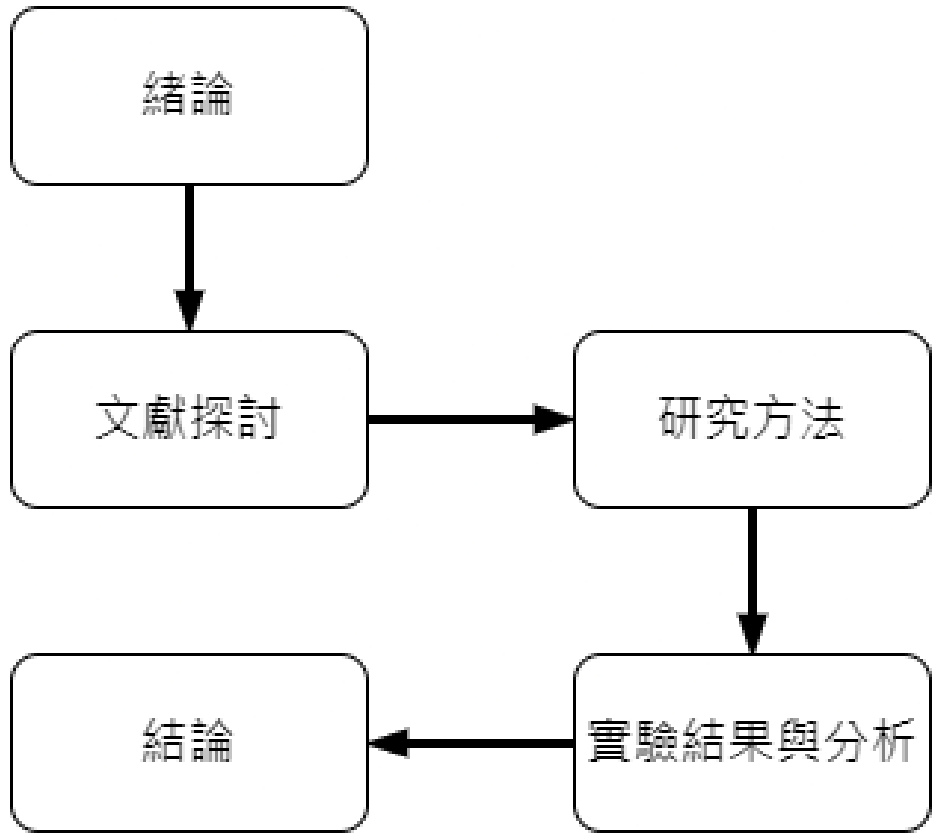

Thesis Structure Diagram

# 2. Literature Review

## 2.1 Blurred License Plates

In the academic context, a "blurred license plate" is defined as a license plate that cannot be recognized by either systems or the human eye [5]. However, there are many reasons for this occurrence, so each study explores different aspects of the issue. Below are three common research directions and related studies:

1. Motion Blur: Some license plate recognition systems often capture images of fast-moving vehicles. Due to the rapid movement of the vehicles, the license plate images become blurred, making it difficult for the system to recognize them correctly. [6] proposes a new system that integrates deblurring and recognition in a closed loop, utilizing the characters and patterns on the license plate as prior information. The process only stops when reliable recognition results are obtained from the deblurred images. Experiments have shown that this system has a higher recognition accuracy on blurred license plates compared to other recognition systems, reaching an accuracy of 86.89%.

2. Low Resolution: This refers to license plate images that are of poor quality or have been compressed by the system, resulting in a resolution too low for accurate recognition. [7] focuses on 4x2 characters, using Bayesian Discrimination to predict blurred license plate characters, presenting the results in a list of possible candidates. [8] employs super-resolution techniques along with single-character segmentation and recognition, combined with a convolutional neural network (CNN) based on VGG-net, achieving an 87.72% accuracy rate in character recognition. [9] combines super-resolution with a perspective distortion correction algorithm, improving the license plate recognition accuracy by 8.8% compared to the original image.

3. External Environmental Factors: This refers to situations where the license plate becomes unrecognizable due to weather conditions (such as snow, rain, fog, etc.), damage to the plate, or obstruction by external elements. [10] addresses license plate images affected by haze, developing a Generative Adversarial Network (GAN) named MPGAN, which can eliminate haze from the images. Besides achieving higher Peak Signal-to-Noise Ratio (PSNR) and Structural Similarity Index (SSIM) compared to other studies, when integrated with a license plate recognition system, the accuracy of license plate recognition also reached 93.9%.

In Taiwan, intersection surveillance cameras mainly come in three different resolutions: 300,000 pixels (640x480), 1 million pixels (1280x720), and 2 million pixels (1920x1080). However, due to budget constraints, many areas still use lower-resolution cameras of either 300,000 or 1 million pixels. Taking Taipei City as an example, while there are about 18,000 intersection cameras, most of them remain at 1 million pixels due to budget proposals not being passed [11]. Even though 1 million pixel cameras are much clearer than 300,000 pixel ones, the images still tend to be quite blurry when capturing license plates from a distance. Moreover, even with an upgrade to 2 million pixel surveillance cameras, the accuracy of license plate recognition only reaches 61.5%[12]( Figure 2)。Additionally, due to insufficient bandwidth, data often needs to be compressed during transmission, which further reduces the resolution of the images, making them even more blurry. Therefore, this study focuses on addressing the issue of blurred license plates caused by low resolution of surveillance cameras or system compression. It aims to resolve the challenge of these images being too low in resolution for clear recognition by systems or the human eye. While previous research on low-resolution license plates has achieved good accuracy in single-character recognition, the accuracy significantly drops when recognizing license plate numbers composed of multiple characters. [13] reports that the initial recognition accuracy for the entire license plate is only 46.52%, requiring up to five attempts to achieve an accuracy rate above 80%. And although [12] has significantly increased the speed of recognition, the accuracy rate still remains around 61%.

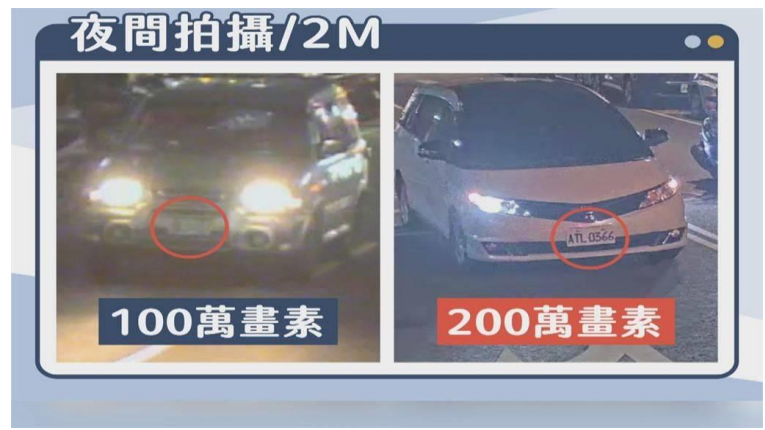

Fig.1 Comparison of Different Pixel Resolution Surveillance Cameras

## 2.2 License Plate Recognition Related Research

With the increasing urban traffic flow and management needs, the demand for license plate recognition technology has been growing. This technology plays a vital role in parking management and traffic monitoring. In the early stages of research, Support Vector Machines (SVM) were used to recognize license plates in parking lots, toll stations, and similar environments, achieving an accuracy rate of 97.2% [14]. By 2006, although license plate recognition technology had advanced to identify plates at various angles and lighting conditions, there was still a need to limit distance and angle to achieve higher recognition accuracy [15] . In recent years, researchers have begun using deep learning to train models for license plate recognition. This shift towards more advanced machine learning techniques has allowed for improved accuracy and versatility in varying conditions. [16] employed YOLOv2 and deep learning to achieve fast and accurate license plate recognition. [17] used YOLOv4, enabling the system to recognize moving license plates even under hardware constraints. [18] focused on improving license plate recognition accuracy in open environments, where issues are caused by lighting conditions. [19] utilized two Convolutional Neural Networks (CNNs), VertexNet for detecting license plates and SCR-Net for license plate recognition, achieving high accuracy rates. [20] combined CNNs with Bidirectional Long Short-Term Memory networks to accurately recognize various types of Malaysian license plates. These studies demonstrate the evolving complexity and effectiveness of machine learning techniques in license plate recognition.

## 2.3 Super-Resolution Related Research

Super-resolution broadly refers to a set of techniques developed to overcome the limitations of sensor and optical manufacturing technologies, enabling the enhancement of low-resolution images or videos to higher resolutions [21]. Its applications are diverse, including surveillance systems, imaging platforms, and more. Super-resolution techniques can be categorized into three types: interpolation-based, reconstruction-based, and learning-based methods. Each of these will be described below.

1. Interpolation-Based Method: This method employs various interpolation techniques (like bilinear, bicubic, or higher-order interpolation) to enhance the resolution of images. [22] The approach involves decomposing the image into appropriate subspaces, performing interpolation on each subspace, and then converting the interpolated values back into the image domain. This technique is typically simpler and less computationally intensive, but it might not always provide the most accurate or detailed results, especially in complex imaging scenarios. During the interpolation process, various optical and structural characteristics of the image can be better preserved, such as the three-dimensional shape of objects, regional uniformity, and local variations in scene reflectance. [23] introduced a novel interpolation method that utilizes multi-surface fitting to fully leverage spatial structural information, achieving super-resolution image reconstruction. This method effectively retains details in the reconstructed high-resolution images by fusing multiple sample values across several surfaces in a maximum posterior probability manner, without needing prior assumptions about the image. This approach is particularly advantageous for maintaining the integrity of complex image structures and fine details. [24] implemented a novel transfer learning approach based on cubic interpolation, introducing a deep learning method that employs a three-dimensional enhanced super-resolution generative adversarial network for reconstructing high-resolution turbulent flows from spatially limited data. This approach leverages the capabilities of deep learning to handle complex three-dimensional data, providing significant improvements in reconstructing detailed features in fluid dynamics. [25] explored the single-image super-resolution problem and proposed a new method combining bilinear interpolation with the U-Net neural network. The process begins with enlarging the image using bilinear interpolation, followed by quality enhancement through the neural network. This method, compared to existing approaches, was able to improve reconstruction quality by 1.00 to 6.56%. This combination of traditional interpolation methods with advanced neural networks allows for a balance between initial upscaling and subsequent refinement, leading to more accurate and higher-quality super-resolution results.

2. Reconstruction-Based Method: These techniques typically rely on mathematical models and use optimization algorithms to reconstruct high-resolution images. [26] introduced a Multi-Improvement Residual Network (MIRN) super-resolution reconstruction model, which enhances the details and textures in reconstructed medical images. This method prevents the images from becoming overly smooth after iterations, a common issue in many reconstruction-based approaches. By focusing on maintaining the integrity of fine details and textures, MIRN significantly improves the quality of medical images, which is crucial for accurate diagnostics and analysis. [27] developed a video super-resolution reconstruction algorithm based on deep learning and spatio-temporal feature self-similarity. This algorithm combines external depth-associated mapping learning with internal spatio-temporal non-local self-similarity priors, enhancing the quality and speed of super-resolution. By integrating these elements, the algorithm effectively leverages both external and internal information, leading to more accurate and efficient upscaling of video content. This approach is particularly beneficial for applications where both high-quality image detail and real-time processing are essential. [28] proposed a single-frame character image super-resolution reconstruction method based on wavelet neural networks. This method integrates wavelet threshold denoising with a wavelet neural network reflectance model, aiming to enhance both the resolution and detail retention of images. By leveraging the wavelet transform's ability to handle multi-scale and multi-level image structures, this approach effectively balances between noise reduction and detail preservation, making it particularly suited for character images where clarity and precision are crucial. [29] addressed issues in face image super-resolution reconstruction methods based on convolutional neural networks (CNNs), such as single-scale feature extraction, low feature utilization, and blurred facial textures. They proposed a model that combines CNNs with a self-attention mechanism. This model starts by extracting shallow features of the image using cascaded 3x3 convolutional kernels. Then, it integrates the self-attention mechanism with residual blocks in a deep residual network to extract deep, detailed features of the face. Finally, the model employs skip connections for global fusion of the extracted features, providing a richer set of high-frequency details for face reconstruction. This approach enhances the clarity and detail of facial features in super-resolved images, which is especially important in applications where facial recognition and analysis are involved.

3. Learning-Based Method: This approach utilizes deep learning, particularly Convolutional Neural Networks (CNNs) and Generative Adversarial Networks (GANs), to learn the transformation from low-resolution to high-resolution images. This method leverages the power of neural networks to implicitly understand complex image features and textures, allowing for more sophisticated and accurate upscaling. This approach often results in superior quality images, capturing finer details that traditional methods might miss. [30] introduced a Convolutional Neural Network trained on both spatial and temporal dimensions of video to enhance its spatial resolution. This approach takes advantage of the temporal continuity in videos, using information from multiple frames to improve the quality of individual frames. By incorporating temporal data, the network can more accurately reconstruct high-resolution details, leading to clearer and more consistent video quality. This method is particularly effective for applications such as video surveillance and broadcasting, where both spatial detail and temporal consistency are important. [31] introduced a highly accurate super-resolution method called VDSR, a very deep convolutional neural network that uses 20 weight layers to improve accuracy. By concatenating small filters multiple times in a deep network structure, it effectively utilizes contextual information from large regions of the image. [32] introduced SRCNN, a deep learning method for single-image super-resolution. This method directly learns the end-to-end mapping between low-resolution and high-resolution images, represented by a deep convolutional neural network. It takes a low-resolution image as input and outputs a high-resolution image. [33] improved SRCNN by redesigning the structure in three aspects. First, a deconvolution layer was introduced at the end of the network, allowing direct learning of the mapping from the original low-resolution image (without interpolation) to a high-resolution image. Second, the mapping layer was reformulated by reducing the input feature dimension before the mapping and expanding it back after the mapping. Third, smaller filter sizes were adopted but with an increased number of mapping layers. The proposed model achieved an over 40-fold speed improvement. In 2017, [34] proposed a Generative Adversarial Network for image super-resolution named SRGAN, which produces images with a Mean Opinion Score (MOS) closer to that of original high-resolution images compared to the aforementioned methods.

## 2.4 Real-ESRGAN

In 2017, [34] published SRGAN, a model that can convert low-resolution images into high-resolution ones while preserving the details of the images.

The following is the architecture of SRGAN (Figure 3):

1. Generator: The primary task of the generator is to map low-resolution images to high-resolution images.

    The generator in SRGAN typically includes the following components:

    - Feature Extraction Layer: This layer uses convolutional layers to extract features from the low-resolution image.

    - Residual Blocks: The generator contains a series of residual blocks, each consisting of two convolutional layers and an activation function (usually ReLU or PReLU). The purpose of the residual blocks is to learn the image residuals (i.e., the differences between the high-resolution and low-resolution images), which helps to improve the performance of the model and accelerate the training process.

    - Upsampling Layer: This layer is used to upsample the feature map to a higher resolution. Common upsampling techniques include transposed convolution and pixel shuffle.

    - Output Layer: This is a convolutional layer used to generate the final high-resolution image.

2. Discriminator: The goal of the discriminator is to differentiate between the high-resolution images generated by the generator and real high-resolution images.

    The discriminator in SRGAN is typically a deep convolutional neural network, which includes:

    - Convolutional Layers: Several convolutional layers are used to extract

features from the images.

- Fully Connected Layers: The extracted features are fed into one or more fully connected layers for the final binary classification (real or generated).

3. Loss Function: SRGAN employs a composite loss function, which includes:
    - Content Loss: Typically a Mean Squared Error (MSE) loss, used to ensure that the generated images are close to the real images on a pixel level.
    - Adversarial Loss: Used to make the generated images more visually realistic and natural.

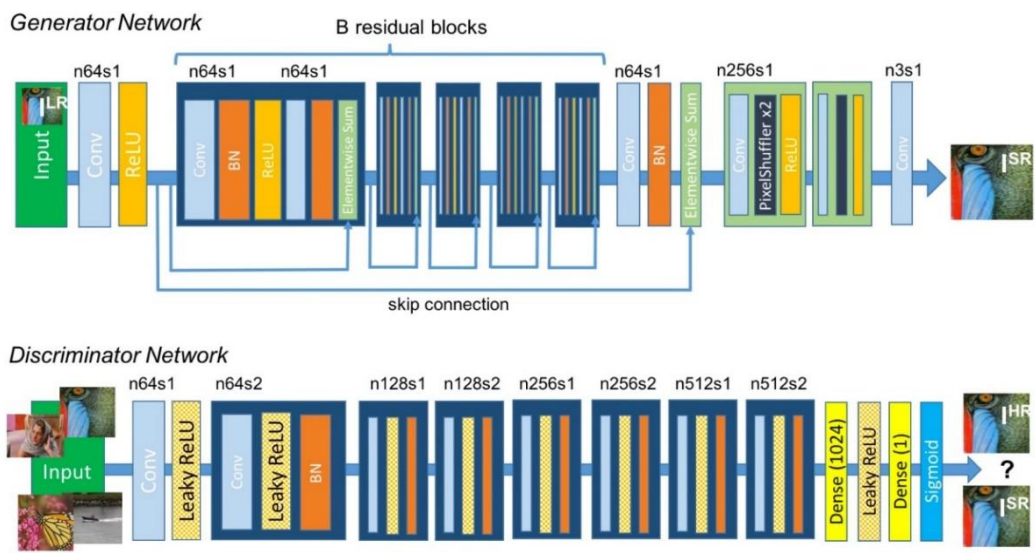

Fig 2. SRGAN Architecture

By 2018, an improved version of SRGAN's architecture named ESRGAN was released [35] . This model produced images with better details compared to those generated by SRGAN (Figure 4).

The following are the differences between ESRGAN and SRGAN:

1. Architecture: Introduced Residual Dense Blocks (RDB) and Residual Channel Attention Blocks (RCAB), further enhancing performance.

2. Loss Function: Added perceptual loss (based on feature extraction using the VGG network) to better capture image details.
3. Training Strategy: Adopted a new training strategy called "progressive refinement," which involves initially training a PSNR-oriented model and then fine-tuning it to improve the visual quality of the generated images.

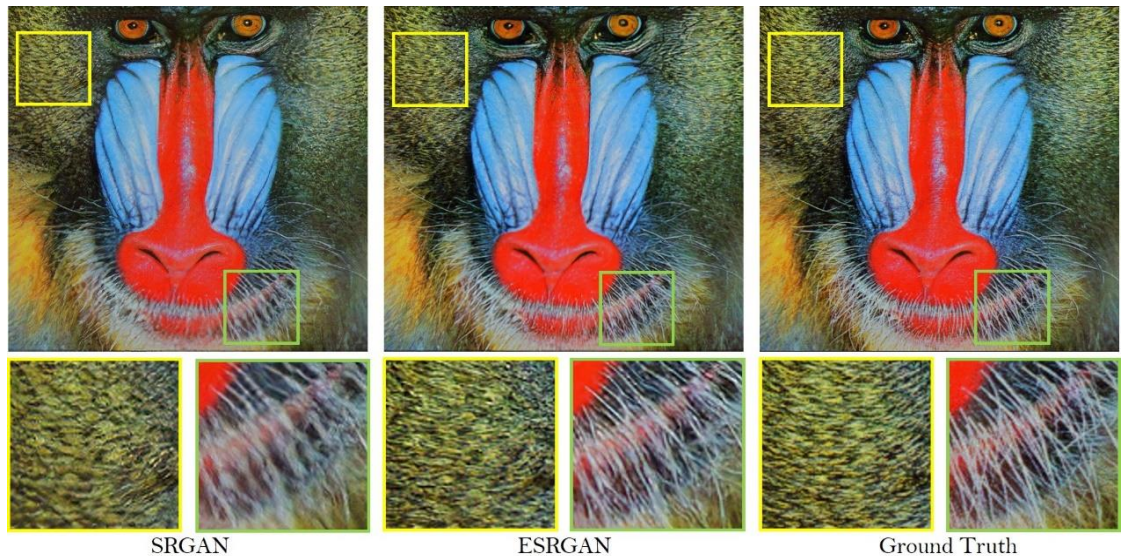

Fig 3. Comparison of SRGAN and ESRGAN

The authors also conducted a detailed comparison of ESRGAN with several other advanced super-resolution models in their study. The results showed that although ESRGAN slightly lagged behind in terms of running speed and computational efficiency, it significantly outperformed the other models in enhancing image details and improving the realism of the images. This difference is primarily attributed to the deep residual networks used by ESRGAN for generating details, allowing the model to more accurately reconstruct high-frequency details, thereby enhancing the clarity and realism of the images. Additionally, ESRGAN employs adversarial loss during training, further enhancing the naturalness and richness of details in the generated images. The combination of these techniques gives ESRGAN a significant advantage in reproducing fine textures and edge details, as evident in Figure 5.

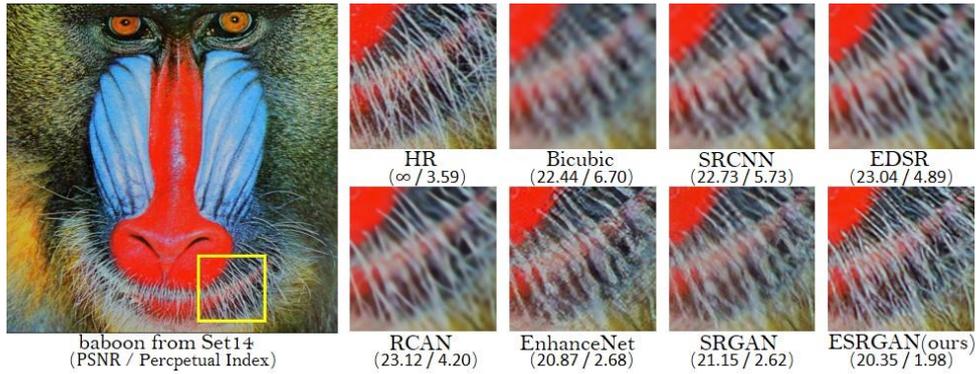

Fig 4. Comparison of ESRGAN with Other Models

Subsequently, the authors of ESRGAN further improved the model, naming the new version Real-ESRGAN [36] .

The following points highlight the improvements in Real-ESRGAN:

1. Developed a new method for constructing datasets, using advanced processing techniques to increase the complexity of downgraded images.
2. Introduced a Sinc filter during dataset construction, addressing the issues of ringing and overshooting in images.
3. Replaced the VGG discriminator used in the original ESRGAN with a U-Net discriminator, enhancing adversarial learning on image details.
4. Incorporated spectral normalization to stabilize the training, which can be unstable due to the complexity of the dataset and the use of the U-Net discriminator.

## 2.5 A-ESRGAN

In 2023, [37] proposed an improved model based on the ESRGAN architecture (Figure 6), named A-ESRGAN. A key feature of A-ESRGAN is its Multi-scale Attention U-Net Discriminator. This discriminator structure includes downsampling encoding modules, upsampling decoding modules, and multiple attention blocks (Figure 7). The model adapts the attention mechanism originally used for semantic segmentation in medical imaging to suit two-dimensional images. This adaptation allows the model to more effectively focus on key details and features within images, thereby improving the quality of super-resolution.

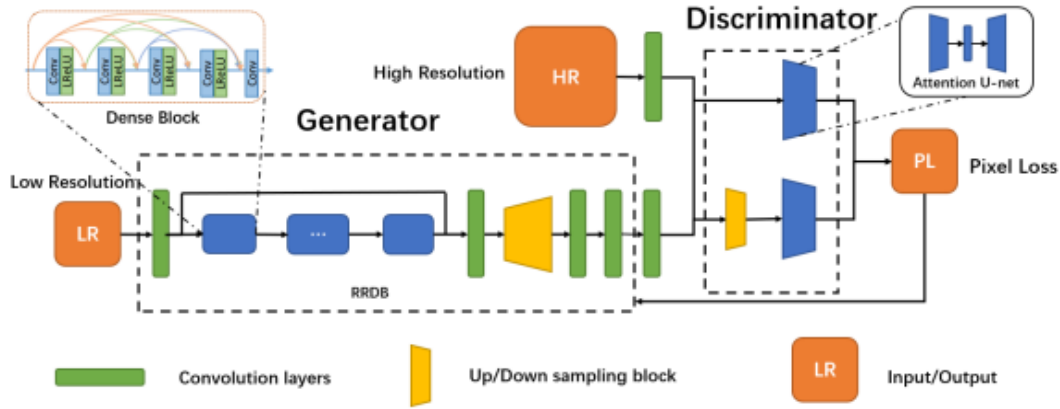

Fig 5. A-ESRGAN Architecture

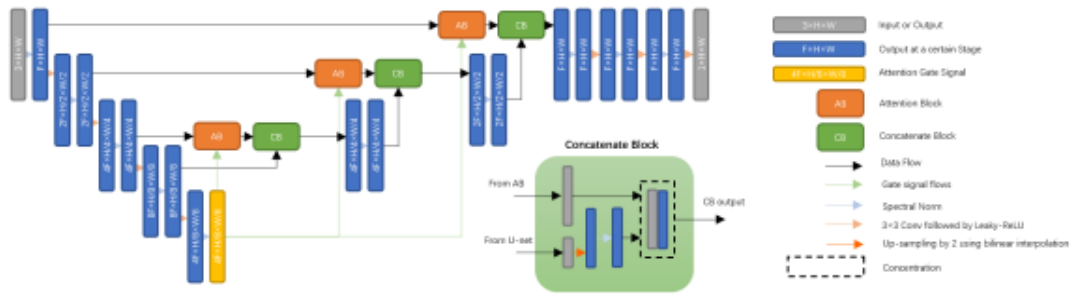

Fig 6. Attention U-Net Discriminator Architecture

## 2.6 StarSRGAN

[38] improved upon Real-ESRGAN and introduced StarSRGAN. The following are the characteristics of StarSRGAN:

1. Star Residual-in-Residual Dense Block (StarRRDB): Compared to the RRDB used in Real-ESRGAN, StarRRDB has a higher capacity and improved ability to reconstruct image details.

2. Multi-scale Attention U-Net Discriminator: The authors combined the multi-scale attention U-Net discriminator from A-ESRGAN [37] with the StarRRDB-based generator, enhancing the ability to recognize fine details in images.

3. Adaptive Degradation Model: This model employs a more effective degradation model that better simulates real-world image degradation.
4. Dual Perceptual Loss: Combines ResNet loss and VGG loss to improve the perceptual quality of images.
5. Dropout Degradation Technique: Used to enhance the generalization ability of the model.

Additionally, the authors developed a lightweight version named StarSRGAN Lite (Figure 9). This version not only offers faster processing speed but also maintains a high image quality, capable of upscaling images by up to 7.5 times.

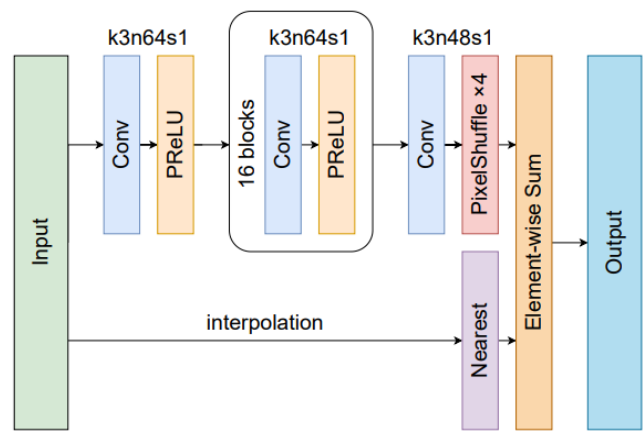

Fig 7. StarSRGAN Architecture

## 2.7 Tesseract OCR

Optical Character Recognition (OCR) is a technology used to identify and convert printed or handwritten text from various image files, such as scanned paper documents, PDF files, or photographs. In 1985, HP Labs developed a new OCR engine named "Tesseract." After more than three decades of development (Figure 9), Tesseract has become one of the most accurate open-source recognition engines available today. Beginning with its fourth edition in 2016, Tesseract shifted from a traditional recognition engine (Legacy Engine) to a recognition engine based on Long Short-Term Memory (LSTM), a deep learning approach (LSTM Engine), which further improved its recognition accuracy. The latest version available is the fifth edition.

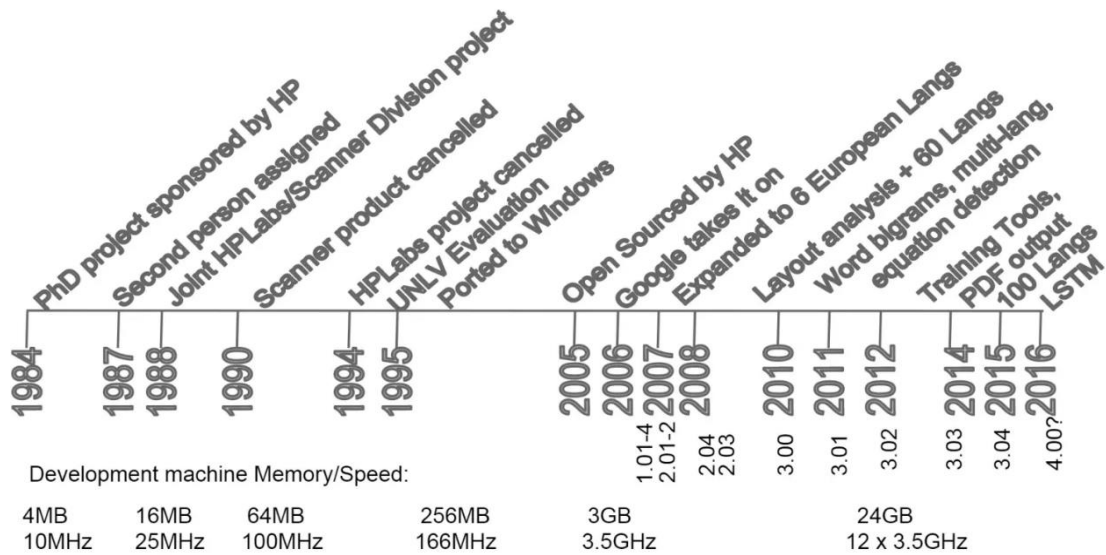

Fig 8. Tesseract Development

Besides accurately recognizing multiple languages[39], Tesseract has also been applied in various other fields. [40] used Tesseract to recognize serial numbers on wine labels, facilitating the tracking and tracing of each wine bottle. Meanwhile, [41] combined OpenCV with Tesseract OCR, enabling researchers to accurately extract license plate numbers from photographs.

The processing workflow of Tesseract can be divided into several main stages:

1. Preprocessing: This includes image binarization, denoising, and correcting image skew. These steps aim to improve the quality of the image, making it more suitable for the subsequent text recognition process.

2. Layout Analysis: Tesseract analyzes the layout of the image to identify text regions, image areas, and other non-text elements. This stage includes the segmentation of lines, words, and characters.

3. Text Recognition: Tesseract uses its built-in feature matching and pattern

recognition algorithms to recognize individual characters. This involves converting character images into feature vectors and then matching them with features in the training data.

4. Post-Processing: This stage includes contextual analysis and error correction. Tesseract uses dictionaries and language models to improve the accuracy of recognition results and correct common recognition errors.

5. Output Formatting: Finally, Tesseract outputs the recognized text in a specified format, such as plain text, HTML, or PDF.

# 3. Research Methodology

In this study, Real-ESRGAN, A-ESRGAN, and StarSRGAN were fine-tuned. The dataset used for training includes the "AOPL" public license plate dataset established by Professor Ji-Sheng Xu of the Department of Mechanical Engineering at National Taiwan University of Science and Technology, as well as images captured by a dashcam. Initially, the data was preprocessed by manually extracting license plates from the photos and outputting the original image as well as versions of the license plate photos scaled down by factors of 4 and 7.5. During the model training phase, fine-tuning was applied. The hyperparameters of the three models were adjusted to achieve the optimal learning rate. Finally, the trained models were used to output enhanced resolution license plate images, which were then recognized using Tesseract to identify the license plate numbers in the images. Model evaluation was based on recognition results, including Confusion Matrix, Accuracy, and Precision. These metrics were used to compare the performance of different models in license plate recognition. The research framework is illustrated in Figure 10.

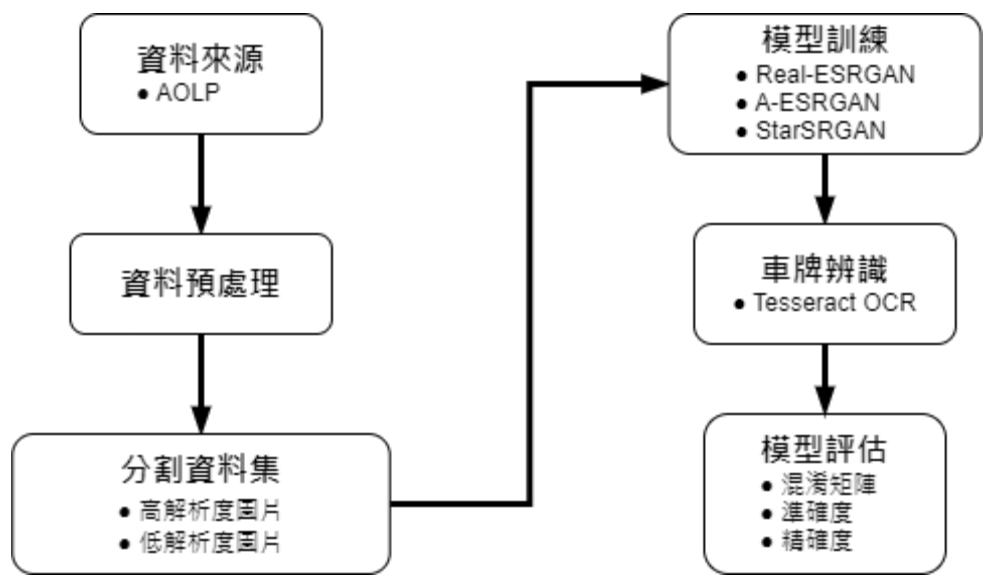

Fig 9. Research Framework Diagram

## 3.1 Data Source

This study collected the "AOLP" public license plate dataset established by Professor Ji-Sheng Xu of the Department of Mechanical Engineering at National Taiwan University of Science and Technology. The entire dataset is divided into three subsets, representing three major application areas:

1. Access Control: This refers to situations where vehicles pass through a fixed channel at reduced speed or come to a complete stop, such as at toll booths or entrances/exits of certain areas. License plate recognition in this context is often used for automated toll collection or controlling vehicle access.

2. Traffic Law Enforcement: This involves vehicles moving at normal or higher speeds but violating traffic laws, such as running red lights or speeding, and then being captured by roadside cameras.

3. Road Patrol: This pertains to cameras mounted on patrol vehicles or handheld by patrol personnel, taking pictures of vehicles from arbitrary angles and distances.

Considering that the number of images in the AOLP dataset might not be sufficient to fully train a super-resolution model, and given that most license plates in the dataset are of the older six-digit format, the model's performance on newer seven-digit plates might not be optimal. To address this issue, we decided to further expand the dataset. We used our own dashcam (2 million pixels) to capture images, and extracted about 1,000 clear images of newer seven-digit license plates from the recorded footage.

The detailed data quantities are shown in Table 1.

Table 1. Dataset Quantity Statistics Table

| Subset Name | Data Quantity |
|---|---|
| Access Control | 681 |
| Traffic Law Enforcement | 757 |
| Road Patrol | 611 |
| Dashcam | 1000 |

## 3.2 Data Preprocessing

During the data preprocessing stage of this study, we used manual labeling to enhance data quality. Specifically, we precisely marked the location of the license plates in each photo and extracted them. The entire extraction process took approximately 15 hours. Subsequently, we copied these extracted license plate photos to a new dataset and used Python to proportionally reduce the size of these photos to one-quarter of their original size. The purpose of this step was to simulate low-resolution surveillance camera images, thereby creating a dataset for super-resolution model training that more closely resembles real-world application scenarios. Through these steps, we were able to effectively evaluate the performance of the three super-resolution models in enhancing the clarity of low-resolution images, as well as their accuracy in recognizing license plates.

## 3.3 Experimental Design

In this study, the dataset is divided into two parts. The first part consists of normal-sized license plate photos, named "High-Resolution Images." The second part includes photos proportionally downscaled, termed "Low-Resolution Images." Apart from uniformly scaling down by a factor of 4, we specifically reduced the size of photos by a factor of 7.5 for the StarSRGAN model to test its super-resolution effectiveness. The average size of the license plate images downscaled by a factor of 4 is approximately 30x10x10, as shown in Figure 11. These images, after reduction, are no longer recognizable by Tesseract in their current state. After data processing, we downloaded three pre-trained super-resolution models: RealESRGAN_x4plus.pth, RealESRGAN_x4plus_netD.pth,A_ESRGAN_Multi.pth,and A_ESRGAN_Multi_D.pth, to serve as the foundation for our super-resolution model training. Next, we fine-tuned and trained the three models to adapt them to our dataset and requirements. This involved adjusting the hyperparameters of each model, with some settings shown in the code screenshot in Figure 12. To maximize the accuracy of license plate recognition, we decided to train these three models at their maximum magnification: Real-ESRGAN and A-ESRGAN were set to magnify 4 times, while StarSRGAN was set to magnify 7.5 times. With such training, our super-resolution models effectively improved the clarity of low-resolution images, thereby enhancing the accuracy of license plate recognition. After training, we used Tesseract OCR to recognize the license plate images processed by super-resolution. We chose to use the official Tesseract language pack eng.traineddata to improve recognition accuracy, thereby enabling a more accurate assessment of the performance of the three models.

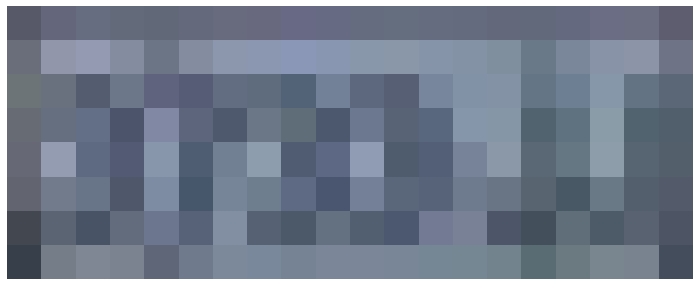

Fig 10. Blurred License Plates

```
# training settings
train:
  ema_decay: 0.999
  optim_g:
    type: Adam
    lr: !!float 1e-4
    weight_decay: 0
    betas: [0.9, 0.99]
  optim_d:
    type: Adam
    lr: !!float 1e-4
    weight_decay: 0
    betas: [0.9, 0.99]

  scheduler:
    type: MultiStepLR
    milestones: [400000]
    gamma: 0.5

  total_iter: 400000
  warmup_iter: -1  # no warm up
```

Fig 11. Hyperparameter Code Screenshot

## 3.4 Model Evaluation

1. **Confusion Matrix**

    Model evaluation is a common tool used to assess the effectiveness of classification models. It illustrates the relationship between actual outcomes and model predictions, often presented in a tabular format, as shown in Table 3. A confusion matrix typically includes the following four parts:

- True Positive(TP)：The actual condition is true, and it is predicted as positive. The prediction matches the actual situation.
- False Positive(FP)：The actual condition is false, but it is predicted as positive. The prediction does not match the actual situation.
- True Negative(TN)：The actual condition is true, but it is predicted as negative. The prediction matches the actual situation.
- False Negative(FN)：The actual condition is false, but it is predicted as negative. The prediction does not match the actual situation.

Table 2. Confusion Matrix

| Actual Condition | Predicted Condition | |
|---|---|---|
|  | Positive | Negative |
| True | True Positive | True Negative |
| False | False Positive | False Negative |

2. **Evaluation Metrics**

    In this study, we will use the values derived from the confusion matrix to calculate evaluation metrics such as Accuracy and Precision.

- Accuracy

Accuracy is a fundamental metric for measuring the performance of a classification model. It describes the proportion of correct predictions made by the model relative to the total number of predictions. It reflects the overall predictive ability of the model. The calculation is as follows:

$$Accuracy = \frac{TP+TN}{TP+TN+FP+FN}$$

- Precision

Precision measures the accuracy of the model when it predicts an outcome as Positive. The calculation is as follows:

$$Precision = \frac{TP}{TP+FP}$$

## 3.5 Experimental Environment

This study utilized the Python programming language and a personal computer as the development tools, as shown in Table 3.

Table 3. Computer Hardware Specifications Table

| Name | Specifications |
|------|----------------|
| OS   | Windows 11 |
| CPU  | AMD Ryzen™ 5 3600X |
| GPU  | NVIDIA GeForce RTX™ 2070 6G |
| RAM  | 16GB |